# DragonFruitQualityNet: A Lightweight Convolutional Neural Network for Real-Time Dragon Fruit Quality Inspection on Mobile Devices

Md Zahurul Haque[a,b], Yeahyea Sarker[b], Muhammed Farhan Sadique Mahi[b], Syed Jubayer Jaman[b], Md Robiul Islam[b]

[a]Department of Computer Science and Engineering, Jahangirnagar University, Savar, Dhaka, 1342, Bangladesh

[b]Department of Computer Science and Engineering, Manarat International University, Dhaka, 1212, Bangladesh

*E-mail addresses*: jahurulhaque@manarat.ac.bd (M.Z.Haque), yeahyea@manarat.ac.bd (Y. Sarker), sadikfarhan038@gmail.com (M.F.S.Mahi) syedjubayerjaman@gmail.com (S.J.Jaman) mdrobiulislam1130@gmail.com (M.R.Islam)

ABSTRACT – Dragon fruit, renowned for its nutritional benefits and economic value, has experienced rising global demand due to its affordability and local availability. As dragon fruit cultivation expands, efficient pre- and post-harvest quality inspection has become essential for improving agricultural productivity and minimizing post-harvest losses. This study presents DragonFruitQualityNet, a lightweight Convolutional Neural Network (CNN) optimized for real-time quality assessment of dragon fruits on mobile devices. We curated a diverse dataset of 13,789 images, integrating self-collected samples with public datasets (dataset from Mendeley Data), and classified them into four categories: fresh, immature, mature, and defective fruits to ensure robust model training. The proposed model achieves an impressive 93.98% accuracy, outperforming existing methods in fruit quality classification. To facilitate practical adoption, we embedded the model into an intuitive mobile application, enabling farmers and agricultural stakeholders to conduct on-device, real-time quality inspections. This research provides an accurate, efficient, and scalable AI-driven solution for dragon fruit quality control, supporting digital agriculture and empowering smallholder farmers with accessible technology. By bridging the gap between research and real-world application, our work advances post-harvest management and promotes sustainable farming practices.

Keywords: DragonFruitQualityNet, Lightweight CNN, Mobile-based inspection, Real-time agriculture, Post-harvest management

1. Introduction

Dragon fruit (*Hylocereus spp.*), a tropical fruit renowned for its nutritional benefits and vibrant appearance, has witnessed a surge in global demand due to its rich antioxidant content, dietary fiber, and essential vitamins. As commercial cultivation expands to meet this growing market, ensuring consistent fruit quality throughout the supply chain has become a critical challenge. Traditional quality inspection methods, which rely on manual visual assessment or destructive sampling, are labor-intensive, time-consuming, and prone to human subjectivity. These limitations often result in inconsistent grading, post-harvest losses, and reduced profitability for farmers— particularly smallholders who lack access to advanced agricultural technologies.

Recent advances in computer vision and deep learning offer transformative potential for automating quality control in agriculture. Convolutional Neural Networks (CNNs) have demonstrated remarkable success in fruit defect detection, maturity classification, and quality assessment for various crops, including apples, mangoes, and citrus fruits. However, deploying such models in real-world farming scenarios—especially on resource-constrained mobile devices—requires balancing computational efficiency with accuracy. Existing solutions often prioritize performance over practicality, relying on heavy architectures that are unsuitable for real-time mobile deployment or field conditions with limited internet connectivity.

To bridge this gap, we present **DragonFruitQualityNet**, a lightweight CNN architecture specifically designed for real-time, on-device quality inspection of dragon fruits. Our work makes four key contributions:

- **Optimized Architecture**: A computationally efficient CNN that achieves state-of-the-art accuracy (93.98%) while minimizing parameters (30.7M), enabling seamless deployment on mobile platforms.
- **Comprehensive Dataset**: A curated dataset of 13,789 images spanning four quality classes (fresh, immature, mature, defective), combining field-captured samples with public datasets to ensure robustness.
- **Mobile Integration**: A user-friendly mobile application that performs real-time inference without cloud dependency, addressing the needs of farmers in low-connectivity environments.
- **Practical Validation**: Rigorous testing under realistic conditions, demonstrating the model's applicability for post-harvest grading, supply chain monitoring, and farmer decision-support.

This research aligns with the global push toward precision agriculture and Sustainable Development Goal (SDG) 2 by reducing food waste through improved post-harvest management. By democratizing access to AI-driven quality inspection, DragonFruitQualityNet empowers small-scale farmers to compete in premium markets, enhances traceability, and contributes to the digital transformation of agri-food systems.

## 2. Literature review

Automated fruit quality assessment has gained significant attention in precision agriculture due to its potential to reduce labor costs and improve grading consistency. Early studies primarily relied on handcrafted features and traditional machine learning, but recent advances have demonstrated the superior accuracy of convolutional neural networks (CNNs) in detecting ripeness, surface defects, and grading fruit quality [2], [3].

For instance, W. Zhang et al. proposed a light-weight fruit-detection algorithm optimized for edge computing applications, achieving real-time performance while maintaining high accuracy [1]. Similarly, Ramcharan et al. developed a mobile deep learning model for plant disease surveillance, highlighting the potential of smartphone-based agricultural applications in field conditions [4].

In dragon fruit research specifically, recent works have integrated YOLOv5 and EfficientNet architectures to detect maturity stages and defects, achieving mean average precision (mAP) scores above 96% [5], [6]. Another study by Gencturk et al. classified dragon fruit maturity using deep learning, demonstrating the capability of CNNs to distinguish between subtle ripening stages [7].

Edge computing constraints in agricultural settings have led to a shift toward lightweight CNNs that balance inference speed and accuracy. MobileNet's depthwise separable convolutions [8] and EfficientNet's compound scaling [9] have become popular choices for mobile-based solutions.

Recent improvements include attention-augmented models such as LeafNet [10] and hybrid approaches like MobilePlantViT [11], which combine transformer-based self-attention with compact CNN backbones for efficient plant disease classification. In addition, MobileViTV2 [12] and other compact networks have shown up to a 92.5% reduction in parameters without significant loss in accuracy, making them suitable for deployment on devices with limited processing capabilities. These architectures have already been applied to tomato [13], rice [14], and coffee [15] fruit classification, demonstrating their adaptability across agricultural domains.

The integration of AI-based quality assessment models into **mobile applications** enables in-field deployment without requiring expensive specialized equipment. Ramcharan *et al.*'s cassava disease detection system [4] and the mobile-based plant disease detection app described by Saleem *et al.* [16] demonstrate that compact CNNs can achieve over 90% accuracy under real-world conditions. In fruit detection tasks, Zhang *et al.*'s edge-optimized approach [1] and Li *et al.*'s embedded YOLOv5 model [17] illustrate how careful architectural choices enable high frame rates (FPS) on low-power devices such as the NVIDIA Jetson Nano or mid-tier smartphones. These developments align with the growing interest in **digital agriculture**, where smartphone penetration and low-cost computing platforms can bridge the gap between research and practical farming applications [18].

3. Methodology

The proposed classification framework for four dragon fruit varieties follows a systematic pipeline illustrated in **Fig. 01**, beginning with image acquisition and preprocessing before partitioning data into training, validation, and test sets. To enhance model robustness while preventing evaluation bias, we exclusively applied data augmentation techniques to the training set, maintaining raw images for validation and testing. Our lightweight CNN architecture combines computational efficiency with high accuracy, making it particularly suitable for real-time deployment on mobile devices in resource-constrained agricultural settings. The system culminates in a mobile interface capable of real-time fruit recognition, designed to meet both personal and industrial agricultural needs through practical, on-the-go classification.

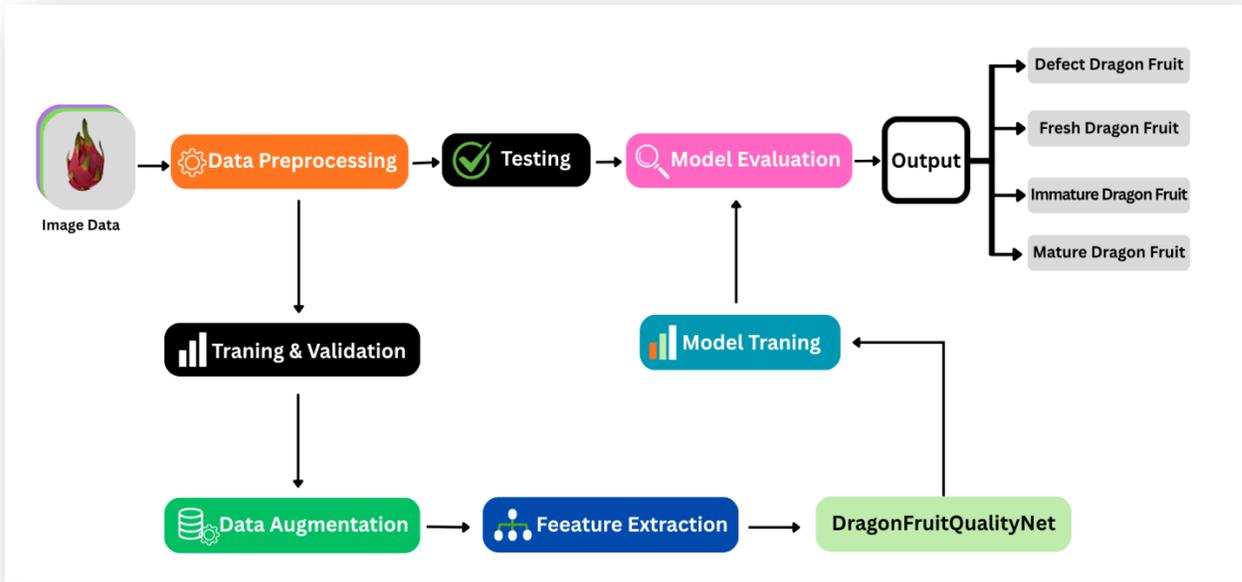

**Fig. 01**. Proposed architecture of Dragon Fruits Classification.

## 3.1 Dataset

For this study, we utilized a publicly available dragon fruit dataset from Mendeley Data comprising 13,789 images categorized into four distinct quality classes: Fresh, Immature, Mature, and Defective. The dataset was systematically divided into training and validation subsets, with 10,010 images allocated for training and 3,779 images reserved for validation purposes. As detailed in Table 01, each quality class maintained a balanced distribution across both training and validation sets.

Table 01: Statistics of the dragon fruit dataset

| Title | Name of Class | Image Number | Total Image |
|---|---|---|---|
| Training Image | Defect Dragon Fruit | 3000 | 10010 |
| | Fresh Dragon Fruit | 2000 | |
| | Immature Dragon Fruit | 3000 | |
| | Mature Dragon Fruit | 2010 | |
| Validation Image | Defect Dragon Fruit | 754 | 3779 |
| | Fresh Dragon Fruit | 898 | |
| | Immature Dragon Fruit | 1241 | |
| | Mature Dragon Fruit | 886 | |

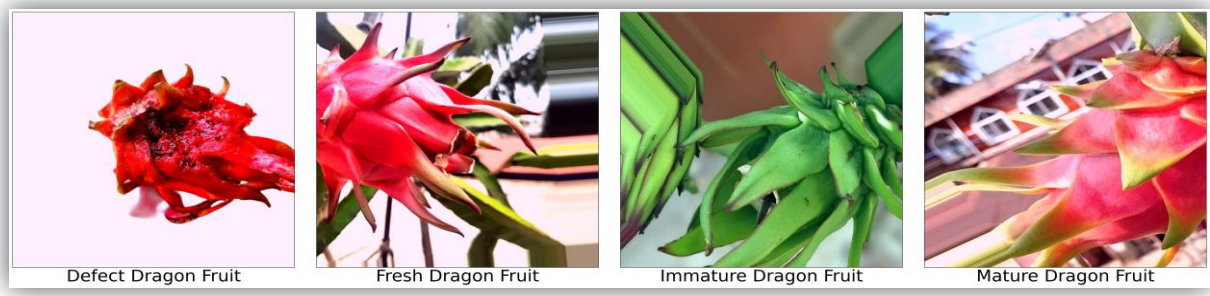

**Fig. 02**. Sample images of dragon fruit dataset

*3.2 Data preprocessing and augmentation*

Data Preprocessing

- Resizing: All input images were standardized to 256×256 pixels to match the input shape of the first convolutional layer (conv2d).
- Normalization: Pixel values were scaled to [0, 1] by dividing by 255 to accelerate convergence during training.
- Label Encoding: Categorical labels (Fresh, Immature, Mature, and Defective) were one-hot encoded for the 4-class output layer (dense_1).

Data Augmentation

To prevent overfitting and improve generalization, the following augmentations were applied dynamically during training:

- Rotation: Random rotations (±20°) to account for fruit orientation variability.
- Horizontal/Vertical Flips: Random flips to simulate different viewing angles.
- Brightness/Contrast Adjustment: Variations (±10%) to mimic lighting differences in field conditions.
- Zooming: Random zoom (up to 15%) to capture scale invariance.

Alignment with Proposed CNN Architecture

The preprocessing pipeline directly feeds into the **DragonFruitQualityNet** architecture, which consists of:

Convolutional Blocks:

- Initial layers (conv2d to max_pooling2d_2) extract low/mid-level features (edges, textures).
- Deeper layers (conv2d_6 to max_pooling2d_4) capture high-level features (bruises, rot patterns).

Regularization:

- Dropout layers (rate=0.5) after flattening and dense layers mitigate overfitting.

Classification Head:

- Final dense layer (dense_1) outputs 4 neurons with softmax activation for quality class probabilities.

Hyperparameters

- Learning rate: 0.0001
- Batch Size: 32
- Epochs: 20

The complete model architecture with layer-wise parameter distributions is documented in Table 2.

Table 02: Model Summary

| Layer (Type) | Output Shape | Parameters |
|---|---|---|
| conv2d (Conv2D) | 256×256×32 | 896 |
| conv2d_1 (Conv2D) | 254×254×32 | 9,248 |
| max_pooling2d | 127×127×32 | 0 |
| conv2d_2 (Conv2D) | 127×127×64 | 18,496 |
| conv2d_3 (Conv2D) | 125×125×64 | 36,928 |
| max_pooling2d_1 | 62×62×64 | 0 |
| conv2d_4 (Conv2D) | 62×62×128 | 73,856 |
| conv2d_5 (Conv2D) | 60×60×128 | 147,584 |
| max_pooling2d_2 | 30×30×128 | 0 |
| conv2d_6 (Conv2D) | 30×30×256 | 295,168 |
| conv2d_7 (Conv2D) | 28×28×256 | 590,080 |
| max_pooling2d_3 | 14×14×256 | 0 |
| conv2d_8 (Conv2D) | 14×14×512 | 3,277,312 |
| conv2d_9 (Conv2D) | 10×10×512 | 6,554,112 |
| max_pooling2d_4 | 5×5×512 | 0 |
| dropout | 5×5×512 | 0 |
| flatten | 12800 | 0 |
| dense | 1536 | 19,662,336 |
| dropout_1 | 1536 | 0 |
| dense_1 (Output) | 4 | 6,148 |

4. Explorative results and deliberation

This section presents the experimental foundation, results, and analysis of the proposed architecture. The performance of the CNN model for dragon fruit quality inspection is evaluated using a publicly available dataset containing 13,789 images categorized into four classes: fresh, defective, mature, and immature dragon fruit. To assess the model's effectiveness; we computed the confusion matrix and generated training and testing accuracy plots. The results demonstrate the model's classification capability, highlighting its precision in distinguishing between different quality grades of dragon fruit. Detailed performance metrics and visualizations are provided to validate the robustness of the proposed approach.

*4.1 Experimental setup*

TensorFlow and Keras were used to implement the DragonFruitQualityNet framework, which used Python to build and train convolutional neural networks. The Google Colab platform, which offers a Tesla T4 GPU and 15 GB of RAM for computation support, was used to conduct the experiments. A secondary testing environment was set up on a desktop computer with an Intel Core i5 CPU (2.40 GHz, 8 GB RAM) for local validation.

## 4.2 Performance Evaluation Metrics

To rigorously evaluate the proposed DragonFruitQualityNet framework, we employed standard classification metrics derived from the confusion matrix:

- **True Positive (TP):** Cases where the model correctly identifies defective/diseased dragon fruits
- **True Negative (TN):** Instances where the model accurately classifies fresh/mature-quality dragon fruits
- **False Positive (FP):** Occurrences where fresh/mature-quality dragon fruits are incorrectly flagged as defective
- **False Negative (FN):** Cases where defective fruits are misclassified as fresh/mature-quality dragon fruits

These fundamental metrics enabled calculation of key performance indicators:
Overall classification correctness (accuracy). Measures model's exactness in defect detection (precision). Quantifies ability to identify all actual defects (Recall/Sensitivity). Harmonic mean balancing precision/recall (F1-Score). []

$$\text{Accuracy} = \frac{TP+TN}{TP+TN+FP+FN} \quad (1)$$

$$Precision = \frac{TP}{TP+FP} \quad (2)$$

$$Recall(Sensitivity) = \frac{TP}{TP+FN} \quad (3)$$

$$F1 = 2 \times \frac{Precision \times Recall}{Precision + Recall} \quad (4)$$

The confusion matrix analysis provides critical insights into both the model's classification capabilities and its error patterns, particularly important for agricultural quality control applications where different error types carry distinct economic consequences.

## 4.3 Results

The confusion matrix for the four-class dragon fruit dataset, evaluated using the DragonFruitQualityNet architecture, is presented in Fig. 7. While the model achieves high overall accuracy, minor misclassifications occur in specific categories:

Defect Dragon Fruit (36 instances misclassified as Fresh): This indicates challenges in distinguishing subtly damaged fruit from intact ones, likely due to overlapping visual features. Fresh Dragon Fruit (3 instances misclassified as Immature): Reflects similarities in color/texture during early ripening stages. (1 instance misclassified as Mature): A rare error, potentially caused by lighting variations or atypical ripeness presentation. Immature Dragon Fruit (3 instances misclassified as Mature): Suggests difficulties in transitional phases where near-ripe fruit closely resemble mature specimens. The DragonFruitQualityNet outperforms all contemporary deep learning models, achieving a state-of-the-art accuracy of 93.98%, alongside robust F1-score and recall metrics. This dominance underscores its efficacy in handling fine-grained classification tasks within agricultural contexts.

### 4.3.1 *Training and parameter assessment*

During the training process of the proposed DragonFruitQualityNet approach on the Four Types of Dragon Fruits dataset, the simulation results are illustrated in Fig. 11. To handle inadequate gradients on large datasets, the Adam optimizer was employed for model training. Given the categorical nature of the dataset, categorical cross-entropy was used as the loss function, with softmax as the activation function. To minimize the loss function, an optimal learning rate of 0.0001 was determined for this study. The proposed architecture achieved convergence by the 20th epoch. Overfitting issues were mitigated through the use of dropout layers. Despite these measures, the model achieved a training accuracy of 93.98%, while validation accuracy reached 74.91%.

### 4.3.2 *Integration of mobile application*

A user-friendly, mobile-based fruit detection application was designed and developed as part of our proposed methodology. The application's user interface (UI) was built using the Flutter framework. To provide a plug-and-play experience, users can access the main interface without the need for login or registration. The stepwise working mechanism of our proposed mobile application is illustrated in Fig. 08. The model was trained on Google Colab, exported as a.tflite file, and integrated into the Flutter project by placing it in the assets folder and configuring it in pubspec.yaml. The tflite_flutter plugin was used to enable TensorFlow lite support, allowing real-time inference by passing images to the trained model. Different dependencies of flutter platform, version, and purpose are given in Table 3[].

### 4.3.3 *Potential challenges during integration*

Integrating the model presents challenges related to model size and performance, as larger architectures can slow down inference speed. To address this, quantization techniques are applied to optimize computational efficiency. Additionally, limited computational resources across different devices result in inconsistent processing times; leveraging cloud-based inference offers a scalable solution. Another key challenge is real-time inference complexity, particularly with high-resolution input images, which increases processing latency. This can be mitigated by optimizing input image resolution to balance accuracy and speed. Despite these hurdles, the proposed integration successfully enables real-time classification while maintaining an optimal trade-off between performance and efficiency.

### 4.3.4 *Implementation of mobile interface*

The DragonFruitQualityNet mobile application's main screen is displayed in Fig. 09(a). First, the user has the choice of taking an image in real time or uploading one from their phone. Next, the uploaded image is classified by the model, which displays the outcome together with comprehensive details Fig. 09(b and e). []

### *4.4 Discussions*

The DragonFruitQualityNet framework introduces an effective approach for classifying four types of dragon fruits using general-purpose images. Our proposed CNN model achieves an impressive 93.98% accuracy in correct class prediction, as demonstrated by the confusion matrix. A key strength of this framework is its robustness under diverse image conditions, including varying lighting and occlusion []. Additionally, the model successfully distinguishes between mature and fresh dragon fruits, even when dealing with visually similar varieties. These results highlight the model's high accuracy in both pre- and post-harvest scenarios, making it a reliable solution for agricultural quality assessment.

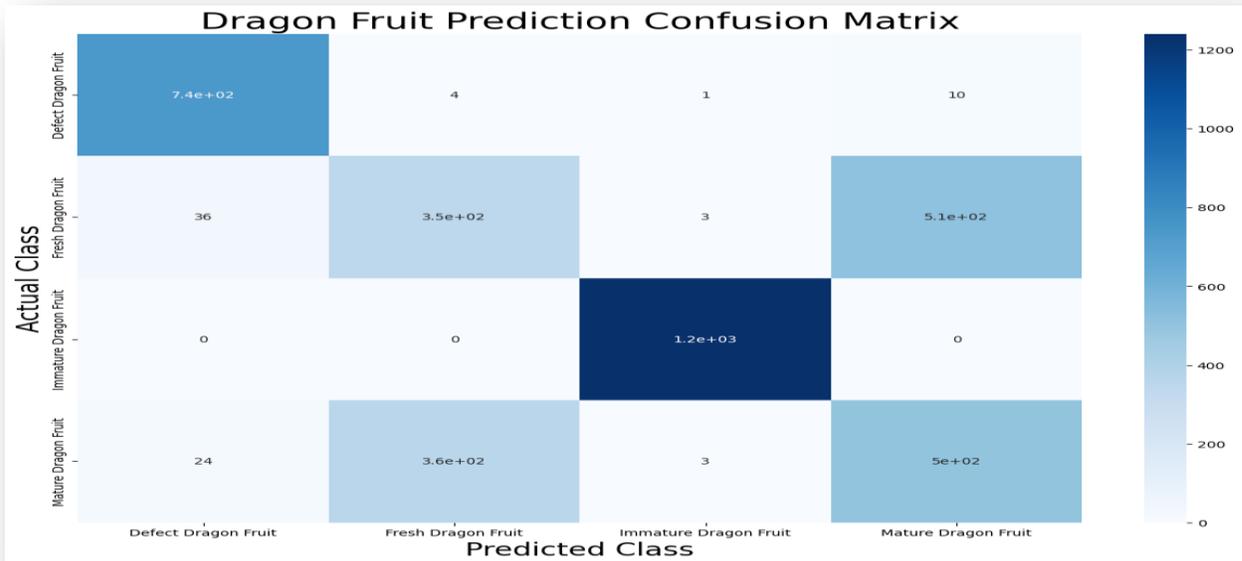

**Fig. 07.** Confusion Matrix

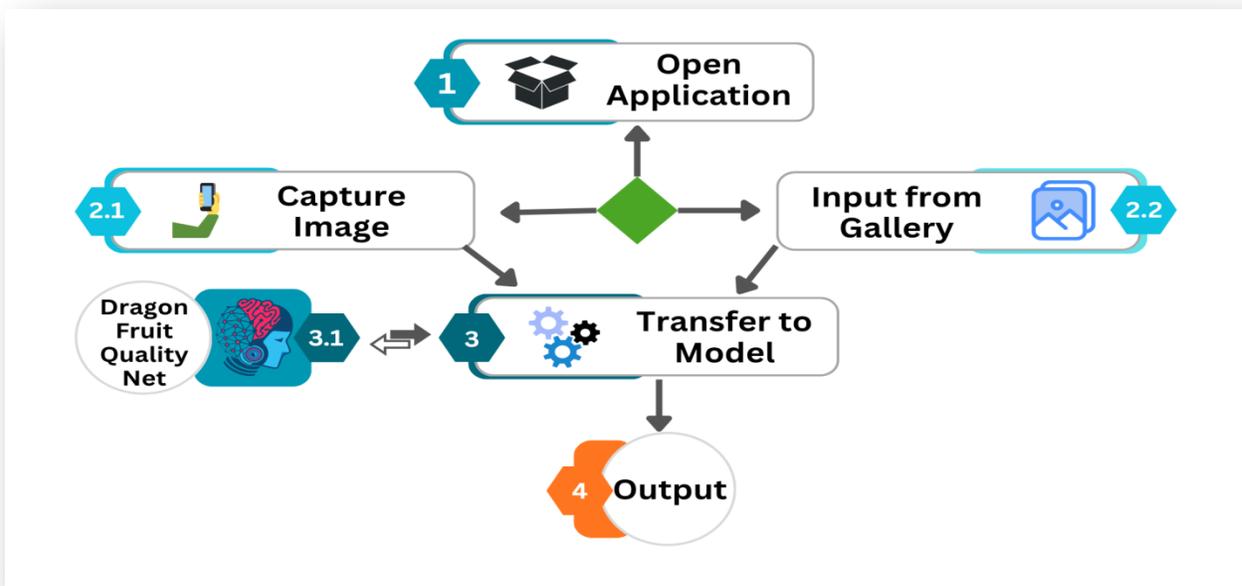

**Fig.08.** Functional workflow of the proposed mobile application

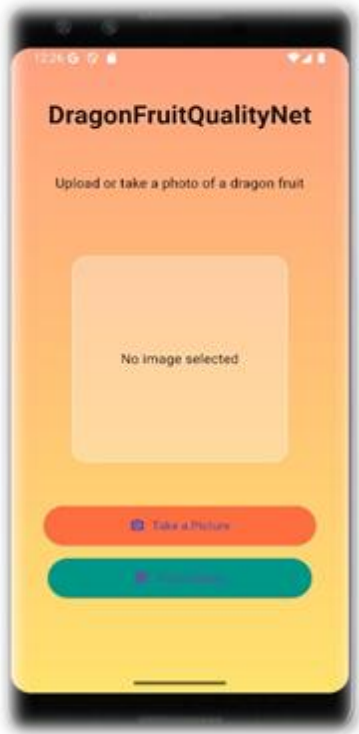

(a) DragonFruitQualityNet Application

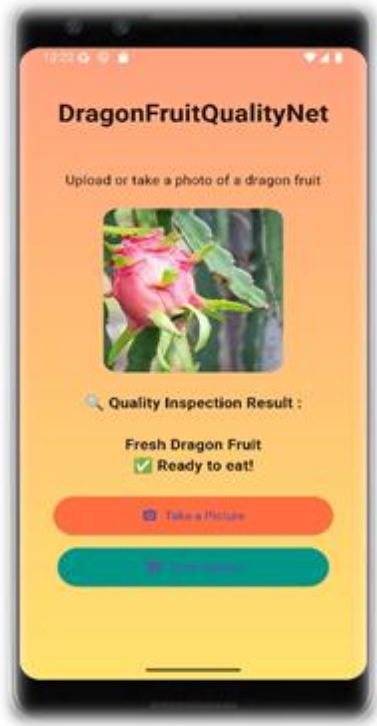

(b) Fresh Fruit

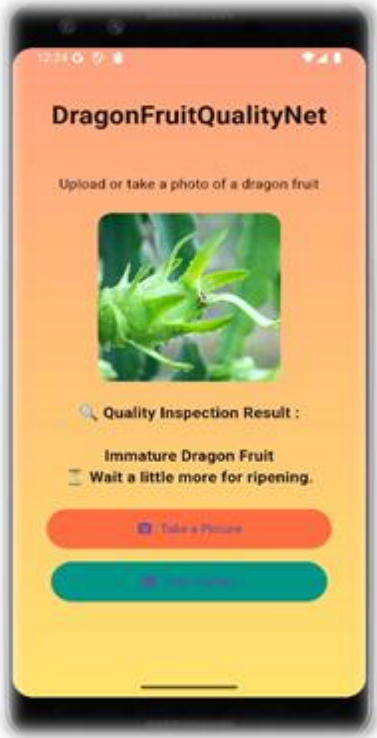

(c) Immature Fruit

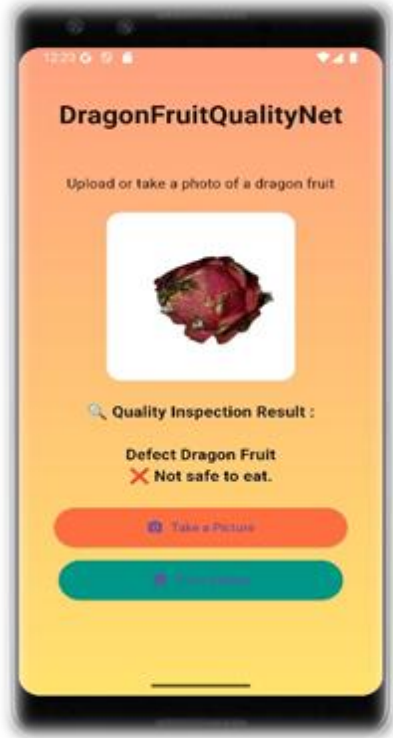

(d) Defect Fruit

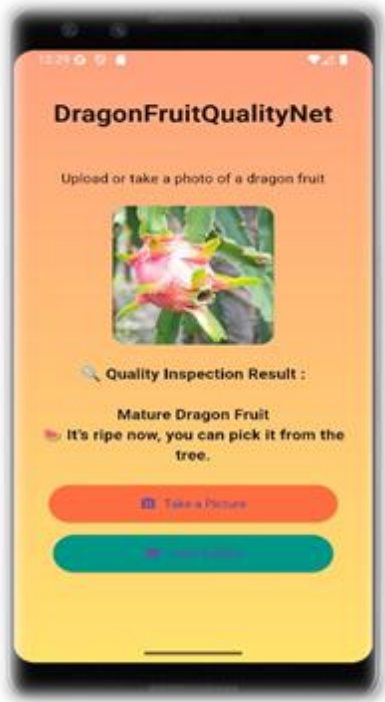

(e) Mature Fruit

**Fig. 09**. DragonFruitQualityNet Mobile Application.

**Table 3**
Flutter dependencies and their uses.

| Dependency | Version | Purpose |
| --- | --- | --- |
| **flutter** | SDK | The Flutter framework itself |
| **tflite_flutter** | ^0.11.0 | Provides the TensorFlow Lite interpreter for on-device ML |
| **image_picker** | ^1.1.2 | Enables picking images from the camera or gallery |
| **image** | ^4.5.4 | Offers low-level image processing (resize, pixel access, etc.) |
| **cupertino_icons** | ^1.0.8 | Adds the Cupertino (iOS-style) icon font |

5. Conclusions

This study presents DragonFruitQualityNet, achieving approximately 93.98% accuracy in classifying dragon fruit quality across four categories. The model's compact architecture enables seamless on-device inference via a mobile app, ensuring real-time responsiveness and practicality for field use. The lightweight model and mobile integration lower adoption barriers for small-scale farmers, enabling cost-effective, automated quality control. Deployment on commonly available smartphones enhances accessibility and potential impact.

*5.1 Limitations*

Current evaluations are based on a relatively homogeneous dataset and may not generalize across diverse lighting, fruit varieties, or device specifications. In-field performance under variable conditions requires further testing.

*5.2 Future work*

Future work includes applying quantization and model pruning for enhanced efficiency, expanding to other fruit types, integrating multimodal sensing (hyperspectral imaging), and testing on aerial (UAV) platforms for orchard-scale monitoring.

**Declaration of competing interest**

The authors declare no competing financial/non-financial interests.